\pdfoutput=1

\documentclass[letterpaper, 10 pt, conference]{ieeeconf}  

\IEEEoverridecommandlockouts

\usepackage{xcolor}
\usepackage[compatibility=false]{caption}
\usepackage{subcaption}
\usepackage{graphicx}
\usepackage[utf8]{inputenc}
\usepackage{adjustbox}
\usepackage{amsmath}
\usepackage{pgffor} 
\title{\LARGE \bf
Walking, Rolling, and Beyond: First-Principles and RL Locomotion on a TARS-Inspired Robot
}
\author{
Aditya Sripada\textsuperscript{1,2}\\
\textit{IEEE RAS Member}\\
adityassripada@gmail.com
\and
Abhishek Warrier\textsuperscript{1}\\
\textit{The Robotics Institute}\\
\textit{Carnegie Mellon University}\\
abhishew@cmu.edu
}

\begin{document}

\maketitle
\thispagestyle{empty}
\pagestyle{empty}

\begingroup
\renewcommand\thefootnote{\arabic{footnote}}
\footnotetext[1]{This work was carried out independently, on the authors’ own time and at their own expense.}
\footnotetext[2]{The author holds an M.S. in Robotics from Carnegie Mellon University and is employed as a Senior Robotics Engineer at Nimble.ai; however, neither institution was formally involved in this study.}
\endgroup

\setlength{\abovecaptionskip}{3pt}
\setlength{\belowcaptionskip}{-4pt}
\begin{abstract}

Robotic locomotion research typically draws from biologically inspired leg designs, yet many human-engineered settings can benefit from nonanthropomorphic forms. TARS3D translates the block-shaped 'TARS' robot from Interstellar into a 0.25 m, 0.99 kg research platform with seven actuated degrees of freedom. The film shows two primary gaits: a bipedal-like walk and a high-speed rolling mode. For TARS3D, we build reduced-order models for each, derive closed-form limit-cycle conditions, and validate the predictions on hardware. Experiments confirm that the robot respects its ±150° hip limits, alternates left-right contacts without interference, and maintains an eight-step hybrid limit cycle in rolling mode. Because each telescopic leg provides four contact corners, the rolling gait is modeled as an eight-spoke double rimless wheel.

The robot’s telescopic leg redundancy implies a far richer gait repertoire than the two limit cycles treated analytically. So, we used deep reinforcement learning (DRL) in simulation to search the unexplored space. We observed that the learned policy can recover the analytic gaits under the right priors and discover novel behaviors as well. Our findings show that TARS3D’s fiction-inspired biotranscending morphology can realize multiple previously unexplored locomotion modes and that further learning-driven search is likely to reveal more. This combination of analytic synthesis and reinforcement learning opens a promising pathway for multimodal robotics.


\end{abstract}
\section{Introduction}
\label{sec:introduction}

\subsection{Background and Motivation}
Research on legged robots has long drawn from biological archetypes such as
bipeds~\cite{Collins2005}, quadrupeds~\cite{raibert2008bigdog}, and
hexapods~\cite{rhex2001hexapod}. Dynamic controllers for platforms such as
ANYmal~\cite{hutter2017anymal} and BigDog~\cite{raibert2008bigdog} have enabled
terrain-adaptive locomotion~\cite{kumar2021terrain}, reinforcing the utility of animal-inspired designs.

However, these forms are not always optimal for human-engineered settings.
Industrial manipulators, vacuum robots, and warehouse shuttles achieve high
efficiency by using task-specific geometries~\cite{Udell_Antropomorphism_2022}.
Recent work has explored \emph{bio-transcending} robots, designs that move beyond
animal morphology to overcome mobility constraints~\cite{Kim_BioTranscending_2023}.
Science fiction often proposes such alternatives, challenging assumptions about
whether anthropomorphism is necessary. This motivates looking beyond traditional animal analogues to forms that exist only in fiction.

\subsection{Science Fiction as a Driver of Robotics Innovation}
Fictional designs have inspired practical robots. The ISS \emph{Astrobee}
free-flyer adapts a cinematic training droid concept for housekeeping
tasks~\cite{Bualat_Astrobee_2019}, while UIUC’s \emph{Ringbot} draws from a
wheel-bike morphology for agile last mile delivery~\cite{Sreenath_Ringbot_2024}.
These cases show that unconventional forms can yield functional solutions.

\begin{figure}[t]
    \centering
    \includegraphics[width=0.35\linewidth]{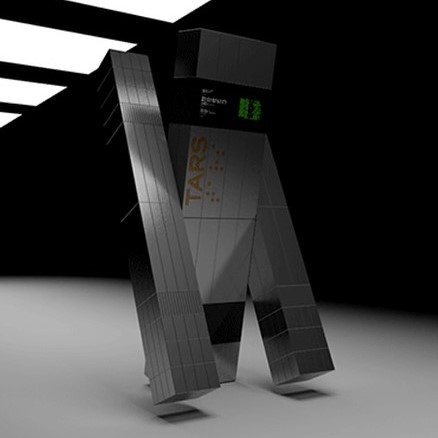}
    \includegraphics[width=0.35\linewidth]{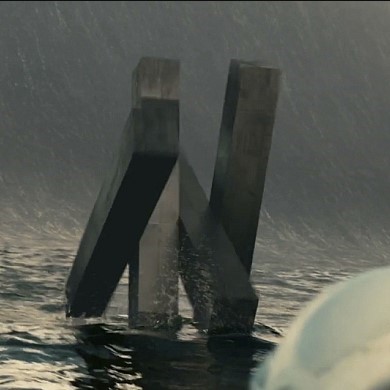}
    \caption{Primary locomotion modes emulated by TARS: 
    (left) bipedal-like gait, (right) rolling gait.}
    \label{fig:three_gaits}
\end{figure}

A notable example is \emph{TARS} from \textit{Interstellar}~\cite{Interstellar2014},
which exhibits two main locomotion modes (Fig.~\ref{fig:three_gaits}):
\begin{itemize}
  \item a \emph{bipedal-like} gait with inner and outer legs moving synchronously, similar to a compass-gait walker
  \item a \emph{rolling} gait where legs reconfigure into a rimless-wheel for high-speed travel.
\end{itemize}

A simple yet kinematically redundant morphology can support multiple
distinct gaits, offering opportunities to study design principles, optimize
locomotor strategies, and expand the operational envelope of mobile robots.
To investigate these possibilities in a real-world setting, we developed
\emph{TARS3D}, a physical platform that translates the cinematic TARS morphology
into an experimentally validated and reconfigurable robot capable of bipedal
and rolling locomotion.

\subsection{Exploring the Design Space Beyond Anthropocentrism}

We analytically derived the two cinematic gaits, bipedal walking and rolling, using closed-form mechanics to identify limit-cycle conditions. To search for additional modes, we trained a PPO~\cite{schulman2017proximalpolicyoptimizationalgorithms} agent in NVIDIA IsaacLab~\cite{mittal2023orbit}. Rolling required structural priors. Fixing the outer hips at $90^\circ$ enabled the policy to replicate the analytic rimless-wheel gait, illustrating how analytic models can seed RL toward high-symmetry motions while RL uncovers novel behaviors.

Previous work shows that optimization can produce unconventional yet effective morphologies, from evolved 'alien' robots~\cite{Lipson_Pollack_2000} to high-performing soft-bodied designs~\cite{Cheney_2014}. Quality-diversity~\cite{Cully_2015} and self-modeling~\cite{Bongard_2006} likewise emphasize task performance over morphological familiarity. TARS3D follows this principle, combining analytically validated walking / rolling gaits with RL-discovered behaviors in an underexplored design space.

\subsection{Contributions}
This work makes four primary contributions:
\begin{itemize}
  \item \textbf{TARS3D}: an additively manufactured platform unifying bipedal-like and rolling gaits in one structure
  \item \textbf{Analytic gait synthesis}: reduced-order models and stability analysis for the two cinematic modes
    \item \textbf{RL-driven gait discovery and adaptation}: a PPO policy, that reproduces the walking \& rolling gaits, uncovers additional novel modes, and maintains stable rolling on uneven terrain up to $\pm0.16\,l_{0}$ uncertainty.

  \item \textbf{Design space expansion}: demonstration that fiction-inspired, non-anthropomorphic morphologies augmented by analytical and learning-based methods can broaden locomotion strategies.
\end{itemize}

\section{System Design}
\label{system_design}

\begin{figure}[!ht]
    \centering
    \begin{subfigure}[t]{\linewidth}
        \centering
        \includegraphics[width=\linewidth]{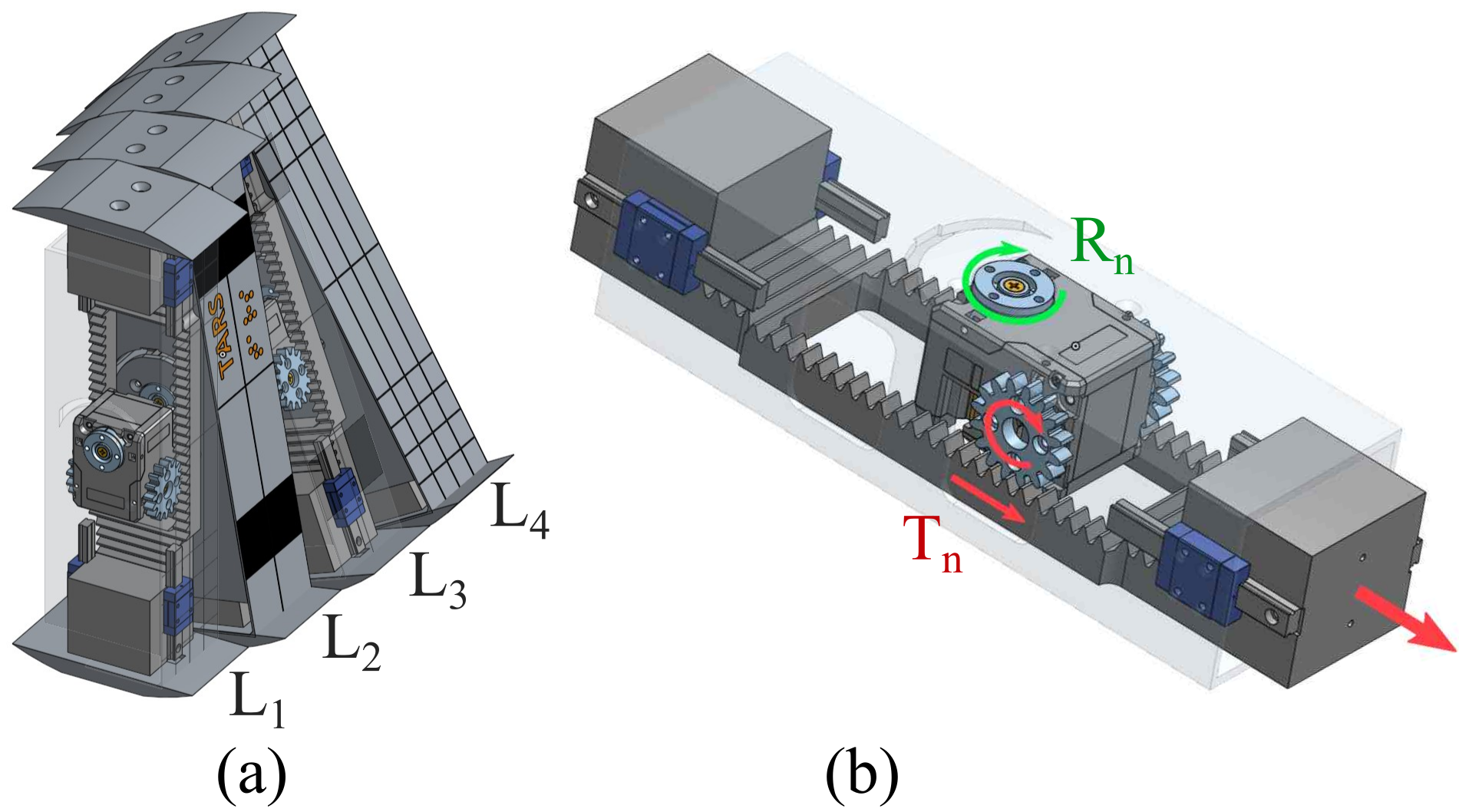}
    \end{subfigure}
    \caption{TARS3D mechanical design: (a) overall CAD model of the 4-leg (${L_1, ..., L_4})$ assembly, (b) leg section showing the rotary hip joint ($R_{n}$) and prismatic leg extension ($T_{n}$)}
    \label{fig:TARS_design}
\end{figure}

To translate the cinematic concept into a functional research platform, we designed
\emph{TARS3D}, an additively manufactured, self-configurable robot that implements the morphology of TARS from \emph{Interstellar}. The robot has eight actuated degrees of freedom: four hip rotations (three active, one passive/free-spinning) and four prismatic leg extensions controlling leg length (Fig.~\ref{fig:TARS_design}a). The 0.248 m tall,
0.99 kg platform uses PLA load-bearing links printed via fused filament fabrication,
with actuators arranged for high power density in a compact envelope.  
Unlike earlier replicas focused on human-robot interaction and basic walking,
TARS3D can (i) reconfigure into a rimless wheel topology for rolling and (ii)
execute both upright walking and rolling gaits using analytically derived,
experimentally validated controllers.

\label{mechanical_structure}
Each leg of TARS3D is driven by a \emph{Robotis Dynamixel~2XL430-W250} with two independently
addressable, back-drivable axes and 12-bit magnetic encoders (0.088$^\circ$ resolution). All joints are commanded in position control mode using the actuator's internal PID loop at 100~Hz over a TTL half-duplex bus. Flat feet from the film were unstable for rolling, as angular momentum is lost at
the first heel strike. We, therefore use curved foot plates with a \(45^{\circ}\) arc
(radius \(0.124~\text{m}\)), discussed further in~\ref{sec:rolling_theory}. The 2 axes of the Dynamixel (Fig. \ref{fig:TARS_design}b) are used for:
\begin{itemize}
  \item \textbf{Hip rotation}: $\pm150^{\circ}$ sagittal motion for gait phasing and reconfiguration. 
  \item \textbf{Telescopic shank}: The rack pinion slider varies in length from $108$ to $134\,\text{mm}$ at $78\,\text{mm/s}$ to allow swing leg clearance and center of gravity (CoG) changes during rolling. 
\end{itemize}

Actuators share a daisy-chained half-duplex TTL bus. Power and data are supplied
through a 3-channel slip ring on the mediolateral axis, enabling 360\textdegree\
rotation without cable twist. Future untethered versions will use distributed
Li-Po cells in the legs to achieve a balanced mass.

\section{Analytical Locomotion Modes: Bipedal Walking and Rolling}
In \emph{Interstellar}, TARS exhibits two distinct gaits, bipedal walking and rolling. We first modeled these analytically, deriving closed-form limit-cycle conditions to assess feasibility on TARS3D. As the robot’s action space is far larger, we later used reinforcement learning to search for additional gaits. We begin with the bipedal gait.
\subsection{Bipedal Gait Model}
\label{biped_gait}
\begin{figure}[!ht]
    \centering
    \includegraphics[width=0.9\linewidth]{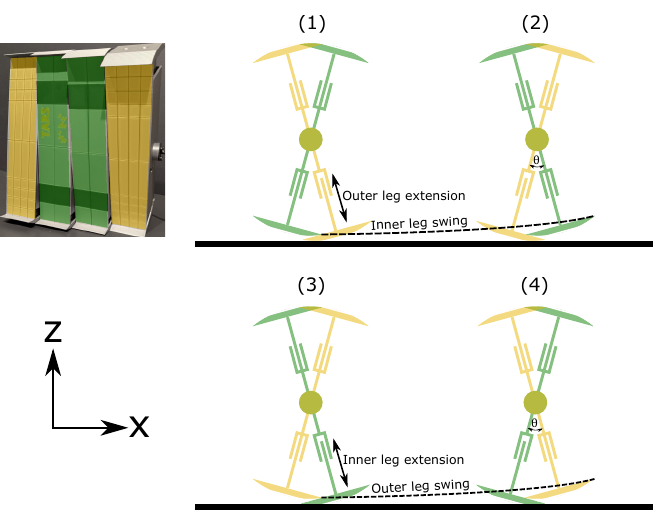} 
    \caption{Phases of bipedal gait of TARS3D}
    \label{fig:biped_art}
\end{figure}

\begin{figure}[!ht]
    \centering

    \makebox[\linewidth][c]{%
      \begin{subfigure}{0.3\linewidth}
          \includegraphics[width=\linewidth]{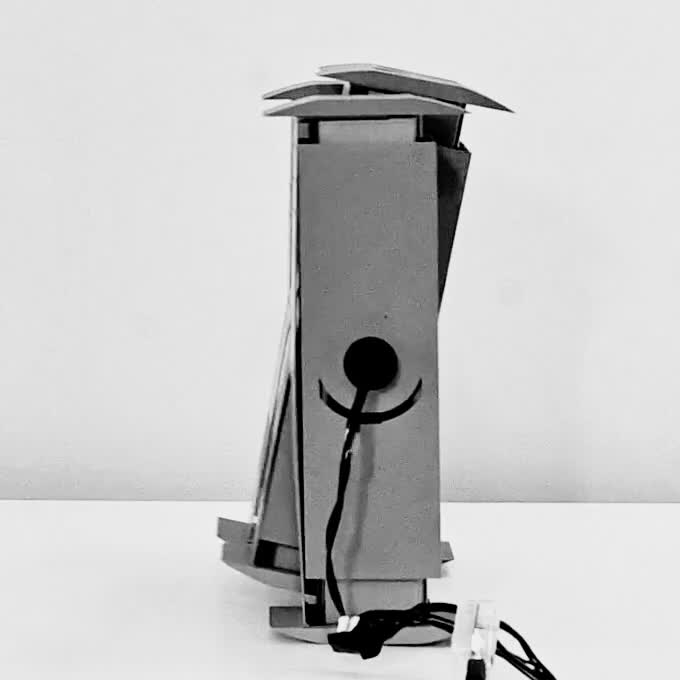}
          \label{subfig:walk_1}
          \subcaption{}
      \end{subfigure}
      \hspace{0.04\linewidth}
      \begin{subfigure}{0.3\linewidth}
          \includegraphics[width=\linewidth]{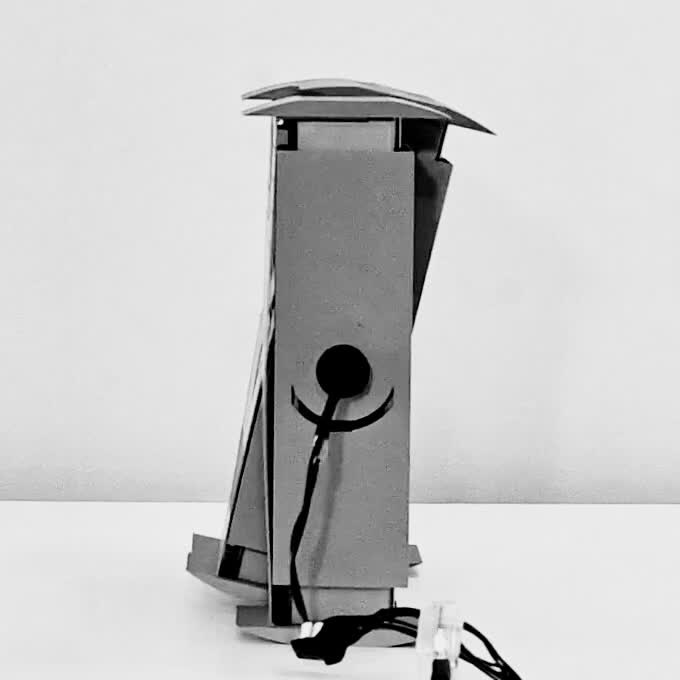}
           \label{subfig:walk_2}
          \subcaption{}
      \end{subfigure}
    }%

    \vspace{4pt}

    \makebox[\linewidth][c]{%
      \begin{subfigure}{0.3\linewidth}
          \includegraphics[width=\linewidth]{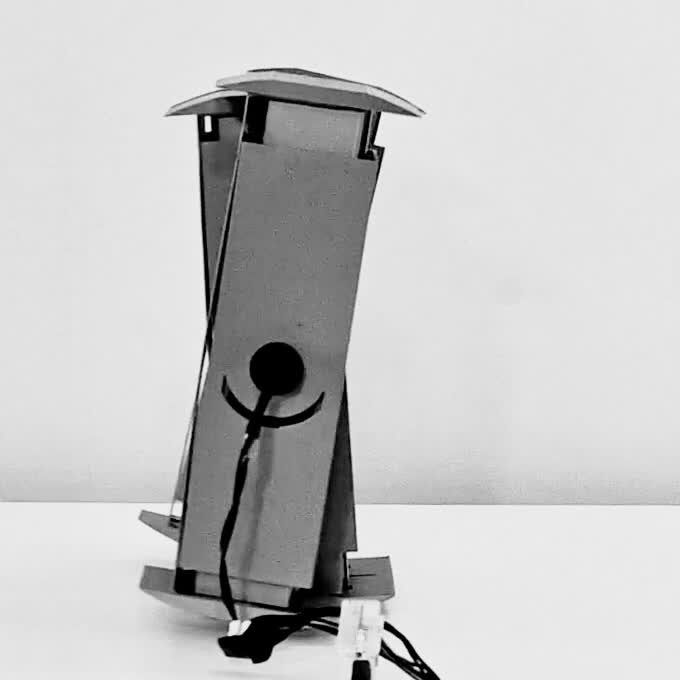}
           \label{subfig:walk_3}
          \subcaption{}
      \end{subfigure}
      \hspace{0.04\linewidth}
      \begin{subfigure}{0.3\linewidth}
          \includegraphics[width=\linewidth]{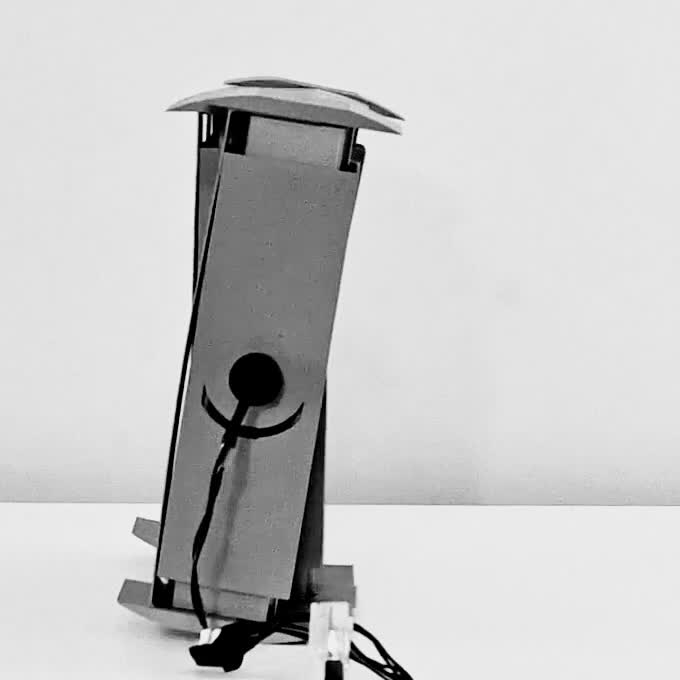}
           \label{subfig:walk_4}
          \subcaption{}
      \end{subfigure}
    }%

    \caption{Four phases of TARS3D bipedal gait}
    \label{fig:tars_four_images}
\end{figure}

The bipedal gait of TARS3D follows the canonical passive curved foot walker sequence~\cite{Garcia1998,Goswami1998,Westervelt2003}:
\begin{enumerate}
  \item \textbf{Outer leg stance/extension}: Rocks on a curved sole, raising the center of mass (CoM) (Fig. \ref{fig:tars_four_images}a).
  \item \textbf{Inner leg swing}: Elevated CoM allows for under-damped swing without scuffing (Fig. \ref{fig:tars_four_images}b).
  \item \textbf{Inner leg stance/extension}: The heel strike triggers leg extension, raising CoM for outer leg clearance (Fig. \ref{fig:tars_four_images}c).
  \item \textbf{Outer leg swing}: Completes the cycle by moving the outer legs forward (Fig. \ref{fig:tars_four_images}d).
\end{enumerate}

\captionsetup[subfigure]{font=small, skip=2pt}

On a virtual slope of angle $\gamma$, gravitational work $mgL\sin\gamma$ offsets collision losses. On level ground, this energy is replaced by a brief stance-leg lengthening impulse~\cite{Asano2008}. Modeling each sagittal plane as a rimless-wheel compass gait~\cite{Kuo2002}, step-to-step energy balance gives:
\begin{equation}
  \Delta E = -\frac{1}{2} m v^{-2}(1-\cos 2\theta),
\end{equation}
with $v^{-}$ the preimpact CoM velocity and $\theta$ the heel-strike angle. $\Delta E=0$ defines a family of limit cycles~\cite{Westervelt2003}. 

The hardware parameters, tuned through the numerical continuation of the simplest walking model~\cite{Garcia1998}, targeted a $35$~mm stride at $0.18$~m/s. The results confirm that curved feet and minimal actuation extend naturally to the morphology of TARS, producing stable bipedal walking, a prerequisite for the more aggressive rolling gait in Section~\ref{sec:rolling_theory}.

\subsection{Rolling Gait Model (Rimless-Wheel Formulation)}
\label{sec:rolling_theory}

\begin{figure}[ht]
  \centering
  \includegraphics[width=0.6\columnwidth,
                   trim={0.1cm 0 0.1cm 0},clip]{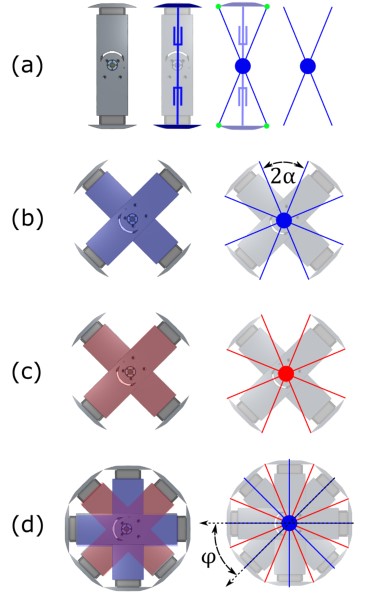}
  \caption{Double rimless-wheel morphology of TARS3D}
  \label{fig:tars_spokes}
\end{figure}

\begin{figure*}[!t]
  \centering
  \adjustbox{max width=\textwidth}{%
    \includegraphics[width=0.8\textwidth,
                     trim={0.1cm 0 0.1cm 0},clip]{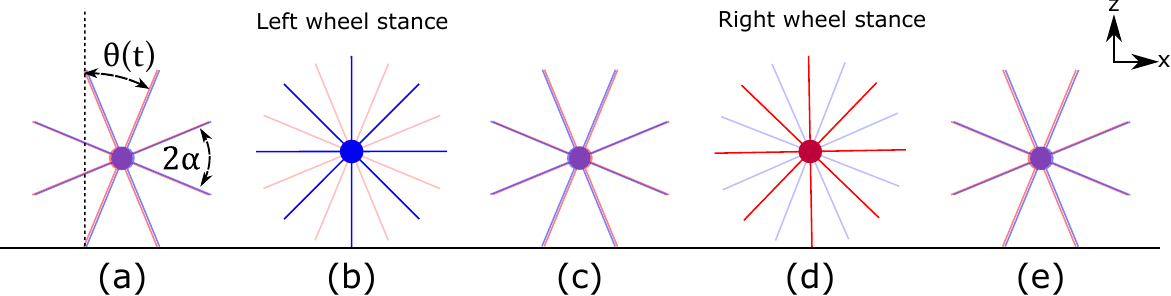}
  }
  \caption{Phases of rolling gait in double rimless-wheel model}
  \label{fig:roll_phases}
\end{figure*}

\begin{figure*}[htbp]
  \centering
  \adjustbox{max width=\textwidth}{%
    \foreach \i in {1,2,...,9}{%
      \includegraphics[width=.11\textwidth]{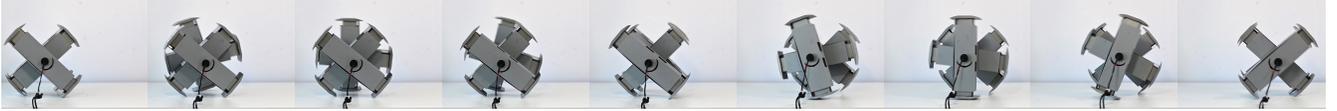}%
    }
  }
  \caption{Phases of rolling gait in \textbf{TARS3D}}
  \label{fig:rolling_gait_banner}
\end{figure*}

Inspired by the efficient passive gait of classical rimless wheels~\cite{McGeer1990,Byl2008}, \textbf{TARS3D} reconfigures its four telescopic legs into a \emph{double-rimless-wheel} (Fig.~\ref{fig:tars_spokes}d). Each \(45^{\circ}\) arc foot plate (\(r = 0.124\,\text{m}\)) yields four equally spaced contacts (Fig. \ref{fig:tars_spokes}a), so one leg acts as four spokes. The pairing of the left legs at $90^{\circ}$ forms an 8-spoke left wheel (\(2\alpha = 45^{\circ}\)) (Fig. \ref{fig:tars_spokes}b), mirrored on the right (Fig. \ref{fig:tars_spokes} c), and the hip joint between them sets the phase \(\phi\) and provides the angular velocity needed to satisfy the vaulting condition.

Rolling alternates stance-leg contact left-right, each wheel advancing a quadrant while the other repositions. Hip travel is limited to \(\pm150^{\circ}\), well below the unlimited rotation of earlier rimless wheel runners~\cite{Bhounsule2014}, so two passive cycles are stitched into an eight-step hybrid limit cycle that completes one body roll per eight ground contacts.

Just before heel strike, the stance leg extends briefly, shifting the CoG forward and injecting an angular impulse \(J^{\ast}\) to offset collision losses and keep \(\dot\theta>\dot\theta_{\min}\). The spokes then resume passive inverted-pendulum motion until the next contact, retaining the efficiency of classical rimless wheels' low controller bandwidth, near-ballistic stance, and favorable energy economy, while enabling high-speed locomotion with joints that cannot rotate indefinitely.

\paragraph*{Dynamics and feasibility}
Assuming inelastic, slip-free impacts, the active wheel behaves as an inverted pendulum.
\begin{equation}
\ddot{\theta} = \frac{g}{l} \sin\theta,
\end{equation}
where $\theta$ is measured vertically. The CoG rises passively from $-\alpha$ to $+\alpha$, converting potential to kinetic energy. At impact, angular momentum conservation about the new contact point gives:
\begin{equation}
\dot{\theta}^+ = \dot{\theta}^- \cos 2\alpha ,
\end{equation}
so each touchdown dissipates a fixed fraction of speed. For the step to be feasible:
\begin{equation}
\dot{\theta}^+ > \theta_{min} = \sqrt{\frac{2g}{l}(1-\cos\alpha)}.
\label{eq:theta_min}
\end{equation}

To compensate for losses, the telescopic joint imparts an angular impulse \(J\):
\begin{equation}
\dot{\theta}^- = \dot{\theta}^+_0 + \frac{J}{m l^2}.
\end{equation}
Combining with the collision map yields the following:
\begin{equation}
\dot{\theta}^+ = \cos 2\alpha \left( \dot{\theta}^+_0 + \frac{J}{m l^2} \right).
\end{equation}
At steady state:
\begin{equation}
J^{\ast} = m l^2 \dot{\theta}^{\ast} \frac{1 - \cos 2\alpha}{\cos 2\alpha}.
\end{equation}

From the angular momentum $H = ml^2\dot{\theta}$, the required leg extension impulse is computed as
\[
\Delta l \approx \frac{J^\ast}{2 m l \dot{\theta}^-},
\]
which, for the hardware parameters ($m = 0.99$~kg, $l = 0.124$~m) and measured pre-impact angular velocity $\dot{\theta}^- \approx 3.1$~rad/s, yields $\Delta l \approx 5$~mm. This matches the capability of the telescopic actuator, which can extend at $\approx 78$~mm/s, implying a $\approx 30$~ms actuation window.

With \(2\alpha=45^\circ\), the model predicts the cost of transport (CoT) as:
\[
v = l \dot{\theta}^{\ast}, \quad
\mathrm{CoT} = \frac{J^\ast \dot{\theta}^\ast}{m g (2 l \sin\alpha)}.
\]
By applying motor torque between the left and right rimless wheels (to overcome the velocity limit in~\eqref{eq:theta_min}) and stance-leg impulses, TARS3D achieved the rolling gait (Fig.~\ref{fig:rolling_gait_banner}). The analytical model assumes perfectly inelastic, slip-free impacts. However, in hardware, brief micro-slip at touchdown (1-3~mm per contact, estimated from high-speed video) slightly alters contact timing but does not disrupt the eight-step cycle. Future work will address micro-slip by incorporating a foot–ground friction model into simulations, complemented by high-friction pads and textured flooring in hardware.

On level ground, TARS3D attains $v\approx0.51$ m/s and $\mathrm{CoT}\approx0.145$, comparable to state-of-the-art rimless-wheel runners~\cite{Bhounsule2014} but without continuous axle rotation. These results confirm that limited-range joints, combined with CoG shift, can deliver efficient, high-speed rolling. The equations here follow canonical rimless-wheel analysis~\cite{Byl2008} with the added CoG-shift mechanism unique to TARS3D.

\section{Exploring Alternative Gaits via Reinforcement Learning}
Deep Reinforcement Learning (DRL) has shown strong results in legged locomotion~\cite{hwangbo2019learning}, enabling discovery of agile gaits with minimal manual design~\cite{li2024reinforcementlearningversatiledynamic}. Given TARS3D's 8-DoF multi-legged morphology, we hypothesize the existence of viable gaits beyond the two analytic modes. We therefore apply DRL, coupled with analytical priors, to explore this space.

\subsection{Setup and Training}
We simulate TARS3D in NVIDIA IsaacLab~\cite{mittal2023orbit}, extending the \texttt{LocomotionEnv} base class used for humanoid and quadruped tasks. The reward combines forward progress (proportional to velocity along the desired heading), an upright bonus to maintain stability, and penalties for energy use, action magnitude, and joint limit violations.

Training uses PPO~\cite{schulman2017proximalpolicyoptimizationalgorithms}, an on-policy algorithm for continuous control. The policy network is a 3-layer MLP ([400, 200, 100] ELU) operating on raw proprioceptive states (joint positions, linear/angular velocities, and related signals), outputting direct joint position commands to bypass lower-level torque control. Additional design choices include a KL-divergence adaptive learning rate scheduler for stability and a slightly elevated entropy bonus to promote exploration. IsaacLab’s GPU-parallelized simulation (2048 environments) allows policy training in under 15 minutes on a single RTX 3070 laptop GPU with 8\,GB VRAM.

\subsection{Learning Results}

\begin{figure}[htbp]
  \centering
  \adjustbox{max width=\columnwidth}{\includegraphics{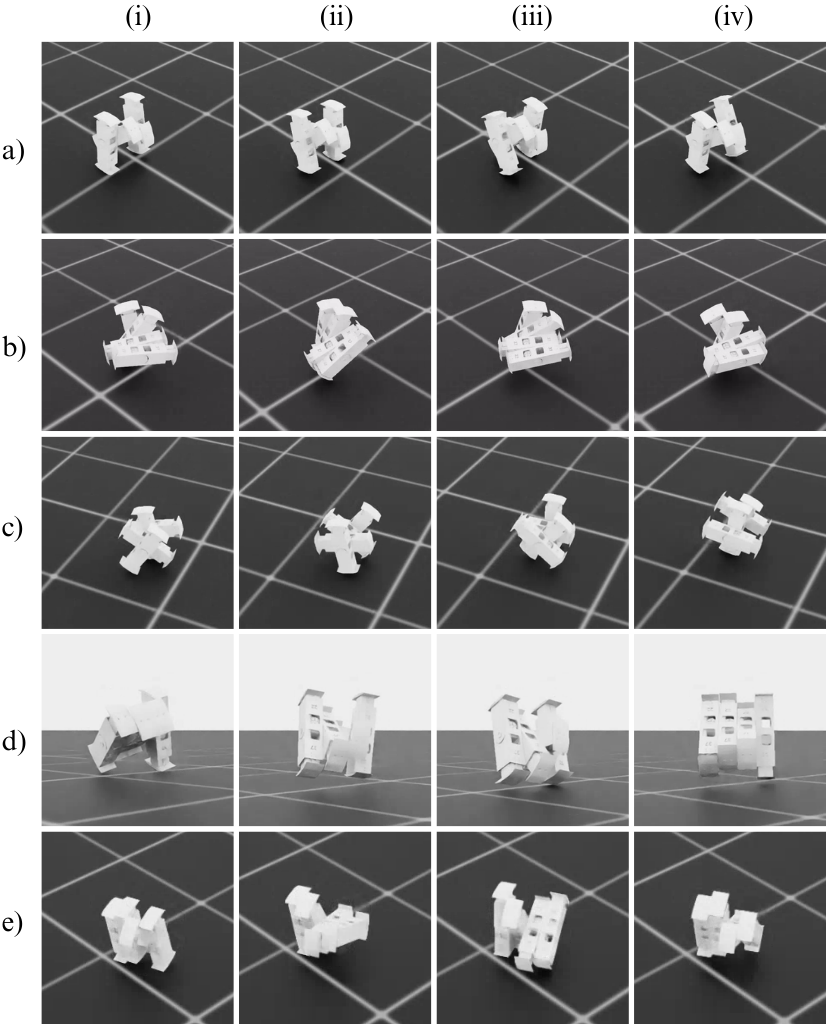}}
  \caption{Behaviors learned through Deep RL}
  \label{fig:rl_res}
\end{figure}

The next section presents an analysis of the center-of-mass (CoM) behavior. We describe the gaits that emerged during training, along with notable failure cases.

\begin{enumerate}
    \item Using the reward function and setup described earlier, the agent learned a bipedal gait (Fig.~\ref{fig:rl_res}a) distinct from the analytical version: the inner legs mostly hover for balance while the outer legs perform the stepping.
    
    \item When the target heading was misaligned with the initial orientation, the agent produced an asymmetric gait (Fig.~\ref{fig:rl_res}b), keeping one side mostly upright while the other oscillated.
    
    \item To encourage rolling, we added an angular velocity reward and locked outer rotational joints at $90^{\circ}$ to form a rimless wheel. This led to successful rolling (Fig.~\ref{fig:rl_res}c) closely matching the analytic solution.
    
    \item Without joint angle priors, reward shaping alone never produced proper rolling. When heading reward dominated, the agent learned a frog-like hop with twisting toward the goal (Fig.~\ref{fig:rl_res}d). When angular velocity reward dominated, it rocked in place to maximize spin without forward progress (Fig.~\ref{fig:rl_res}e).
\end{enumerate}

\subsection{Terrain Adaptation}

We evaluated the robustness of the policy on uneven terrain using Isaac’s \texttt{TerrainGenerator}. Two setups were tested: (a) training directly on randomized terrain, which consistently converged to walking rather than rolling, and (b) transferring a flat-ground rolling policy to randomized terrain. The latter succeeded up to moderate uncertainty but degraded with larger perturbations (Table~\ref{tab:terrain}).

\begin{table}[ht]
  \centering
  \caption{Rolling policy success under terrain uncertainty. $l_{0}$ denotes the unextended leg length.}
  \label{tab:terrain}
  \begin{tabular}{c c c}
    \hline
    Uncertainty & Relative to $l_{0}$ & Success rate \\
    \hline
    $\pm$1.0 cm & $\sim$0.16 $l_{0}$ & Stable rolling \\
    $\pm$1.25--1.5 cm & $\sim$0.20--0.25 $l_{0}$ & $\sim$25\% \\
    $>$1.5 cm & $>$0.25 $l_{0}$ & Failure \\
    \hline
  \end{tabular}
\end{table}

Rolling thus proved highly sensitive to terrain variation, while walking remained a robust fallback. Increasing uncertainty also caused systematic heading drift, probably due to perturbed leg placement redirecting contact forces into turning moments. Similar foot placement-induced yaw effects are well known in running systems~\cite{sripada2021thesis}. Our RL results extend this principle to rolling locomotion, suggesting that foot-placement-aware rewards or corrective feedback could improve robustness on irregular terrain.

\begin{figure}[ht]
  \centering
  \includegraphics[width=0.5\columnwidth]{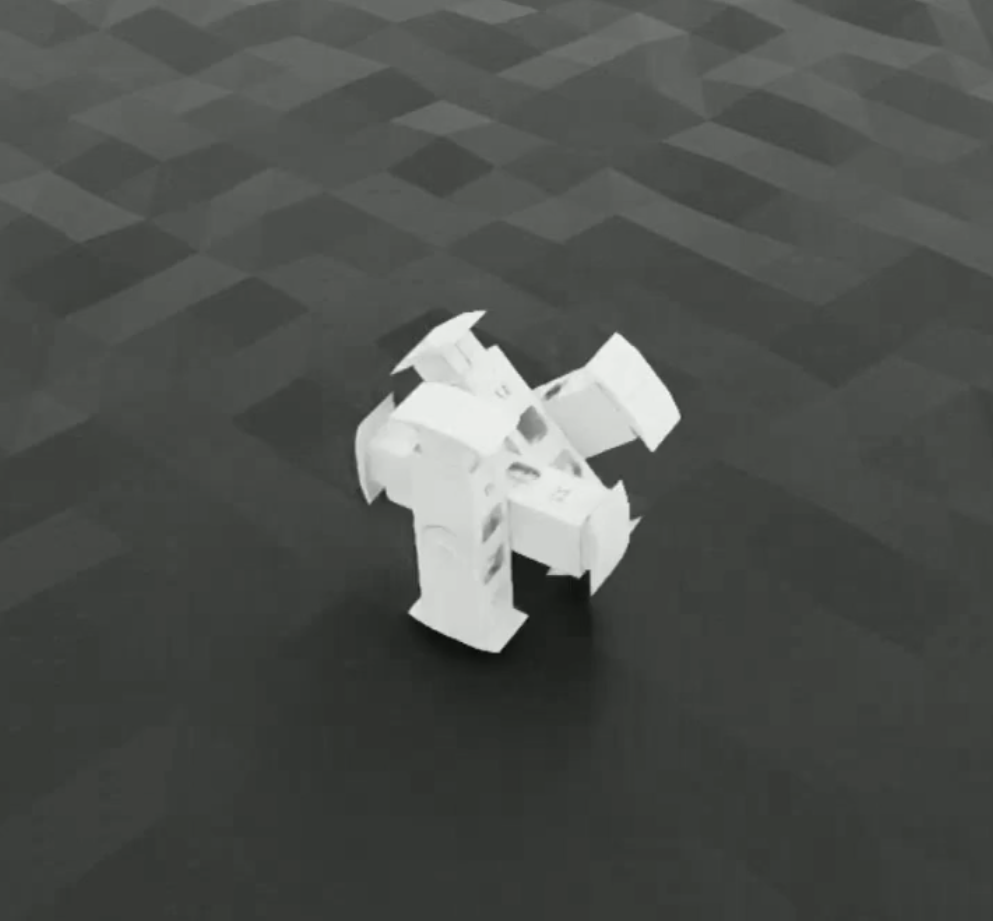}
  \caption{TARS3D rolling on randomized terrain with $\pm 0.16 l_{0}$ uncertainty, generated using Isaac’s \texttt{TerrainGenerator}.}
  \label{fig:terrain_roll}
\end{figure}

Fig.~\ref{fig:terrain_roll} illustrates the robot that performs the rolling gait on randomized terrain with perturbations $\pm 0.16 l_{0}$, a regime in which stable rolling was consistently maintained.

\section{Discussion}

\subsection{Multiple gaits from a non-anthropomorphic morphology}
Our results show that a simple, nonanthropomorphic morphology can realize multiple, fundamentally distinct gaits. The minimally actuated curved foot walker (Sect.~\ref{biped_gait}) reproduces the bipedal-like gait of \emph{Interstellar}, while the double-rimless-wheel model (Sect.~\ref{sec:rolling_theory}) predicts closed-form limit cycles that closely match the rolling gait in hardware. In both cases, TARS3D respects its $\pm150^\circ$ hip limits, alternates left-right contacts without interference, and sustains the hybrid rolling cycle (Fig.~\ref{fig:rolling_gait_banner}).

\subsection{Role of reinforcement learning in gait discovery}
Reinforcement learning (RL) complements this first-principles approach by exploring additional regions of the gait manifold. At low target speeds ($<0.5\,$m/s), PPO discovered novel behaviors, such as diagonal pacing and a “frog hop,” absent from our analytic models. However, for complex motions like rolling, learning failed without strong priors, even with extensive reward tuning. Adding structural priors, joint constraints, posture morphing, and contact alternation shaping led the policy to converge to the analytic rolling gait, achieving $v \approx 0.51\,$m/s and $\mathrm{CoT} = 0.18$.

\begin{figure}[!ht]
  \centering
  \includegraphics[width=0.8\columnwidth]{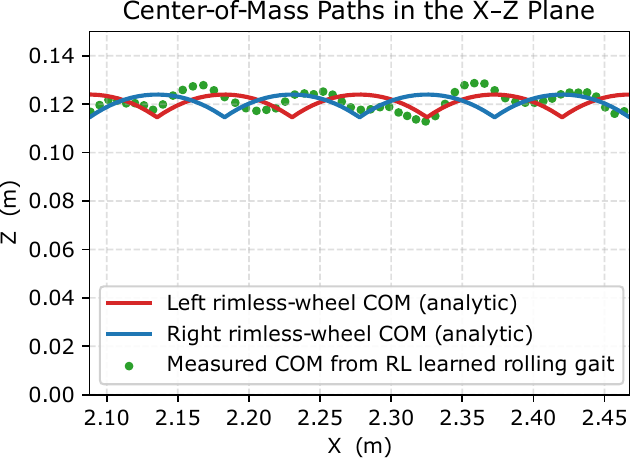}
  \caption{CoM of TARS3D's rolling gait from analytical model vs. RL-learned gait}
  \label{fig:com_plot}
\end{figure}

\subsection{Comparison of learned and analytic rolling gait}
Comparison of the CoM trajectories from the RL-learned rolling gait to the analytic model (Fig.~\ref{fig:com_plot}) shows close overlap, confirming that the learned policy reproduces both the high-level pattern and low-level kinematic constraints. This validates the analytic model and highlights the value of embedding model-based priors in learning.

\subsection{Proposed workflow for multimodal locomotion}
In general, these findings suggest a generalizable workflow for multimodal locomotion research: (1) derive interpretable seed solutions via reduced-order models and (2) use data-driven methods guided by those seeds to explore higher-dimensional control spaces. TARS3D demonstrates that biotranscending morphologies inspired by fiction can validate cinematic gaits and reveal behaviors beyond conventional bipeds or quadrupeds~\cite{Lipson_Pollack_2000,Cheney_2014}.

\subsection{Limitations and future work}
The current prototype is tethered and tested only on flat terrain, with limited optimization for speed and energy efficiency and limited terrain diversity or sim-to-real coverage for learned gaits. Future work will focus on untethered operation with on-board power and communication, robust mode switching between walking and rolling, adaptation to unstructured or compliant surfaces, broader terrain and disturbance testing, and quantitative energetic analysis. Richer priors, such as human demonstrations or motion capture, may further accelerate the learning of complex behaviors. For eventual deployment, the RL policy's position outputs can be directly mapped to Dynamixel position commands via the same PID interface used for analytic gaits, allowing reuse of the existing communication and control stack. Base velocity estimation for policy inputs can be obtained from the onboard IMU, fused with leg odometry via a complementary filter to mitigate drift. Such sensing is not yet integrated on the current tethered prototype.

\subsection{Summary}
This work translates the cinematic TARS morphology into a functional research platform, demonstrating how analytic models and reinforcement learning can jointly enable multimodal, non-anthropomorphic locomotion. Curved-foot walking and eight-spoke rolling were derived in closed form and validated on hardware, while a prior-guided PPO policy reproduced rolling and uncovered novel low-speed gaits. These results highlight fiction-inspired morphology as a viable path to expanding locomotion strategies beyond biomimicry.

\bibliographystyle{IEEEtran}
\bibliography{sn-bibliography}
\end{document}